\documentclass[conference]{IEEEtran}
\usepackage{times}

\usepackage[numbers]{natbib}
\usepackage{multicol}
\usepackage[bookmarks=true]{hyperref}
\usepackage{graphicx}
\usepackage{caption}
\usepackage[bottom]{footmisc}
\usepackage{dblfloatfix}
\usepackage{amsmath}
\usepackage[usenames, dvipsnames]{color}
\usepackage{multirow}
\usepackage[table,xcdraw]{xcolor}
\usepackage{tabularx}
\usepackage[export]{adjustbox}
\usepackage{xspace}
\usepackage{mfirstuc}
\usepackage{wrapfig}
\usepackage{balance}
\graphicspath{ {graphics/} }

\usepackage[normalem]{ulem}
\useunder{\uline}{\ul}{}

\usepackage{hyperref}

\makeatletter
\def\upcase{\expandafter\makeupcase}
\def\makeupcase#1{\uppercase{#1}}

\newcommand{\ingress}{{\small INGRESS}\xspace}
\newcommand{\cmd}[1]{\textsc{#1}\xspace}

\newcommand{\self}{{self-referential}\xspace}
\newcommand{\rel}{{relational}\xspace}
\newcommand{\selfCaps}{{Self-Referential}\xspace}
\newcommand{\relCaps}{{Relational}\xspace}
\newcommand{\snn}{\mbox{S-LSTM}\xspace}
\newcommand{\rnn}{\mbox{R-LSTM}\xspace}
\newcommand{\rset}{\ensuremath{\mathcal{R}}}
\newcommand{\ingressS}{{\small S-INGRESS}\xspace}
\newcommand{\ingressSR}{{\small INGRESS}\xspace}

\renewcommand{\eqref}[1]{(\ref{#1})}

\newcommand{\subfig}[1]{\textit{#1}}

\newcommand{\eg}{\textrm{e.g.}}
\newcommand{\etc}{\textrm{etc.}}
\newcommand{\etal}{\textrm{et~al.}}

\newcommand\prob{\ensuremath{p}}
\DeclareMathOperator*{\argmax}{arg\,max}
\newcommand\given[1][]{\:#1\vert\:}

\usepackage{color}
\definecolor{darkgreen}{rgb}{0,0.60,0.30}
\definecolor{fullred}{rgb}{0.95,.0,.1}
\newcounter{cmt}

\newcommand{\tinylineskip}{\baselineskip=4pt}
\newlength{\citeskipup}
\newlength{\citeskipdown}
\newcommand{\citeinfo}[2]{{\footnotesize \tinylineskip
 \vspace*{\citeskipup}\noindent {\sc #1} \\
 {#2} \vspace*{\citeskipdown}}}

\begin{document}

\title{Interactive Visual Grounding of Referring Expressions for Human-Robot Interaction}


\author{\authorblockN{Mohit Shridhar}
\authorblockA{
School of Computing\\
National University of Singapore\\
Email: mohit@u.nus.edu}
\and
\authorblockN{David Hsu}
\authorblockA{School of Computing\\
National University of Singapore\\
Email: dyhsu@comp.nus.edu.sg}}


%

\twocolumn[{%
\renewcommand\twocolumn[1][]{#1}%
\maketitle

\setlength{\citeskipup}{-2.5truein}
\setlength{\citeskipdown}{2.2truein}
\citeinfo{appeared in}{\textit{Proc. Robotics: Science \& Systems (RSS), 2018}}

\vspace{-20pt}

\begin{center}
    \centering
    \includegraphics[width=1.0\textwidth]{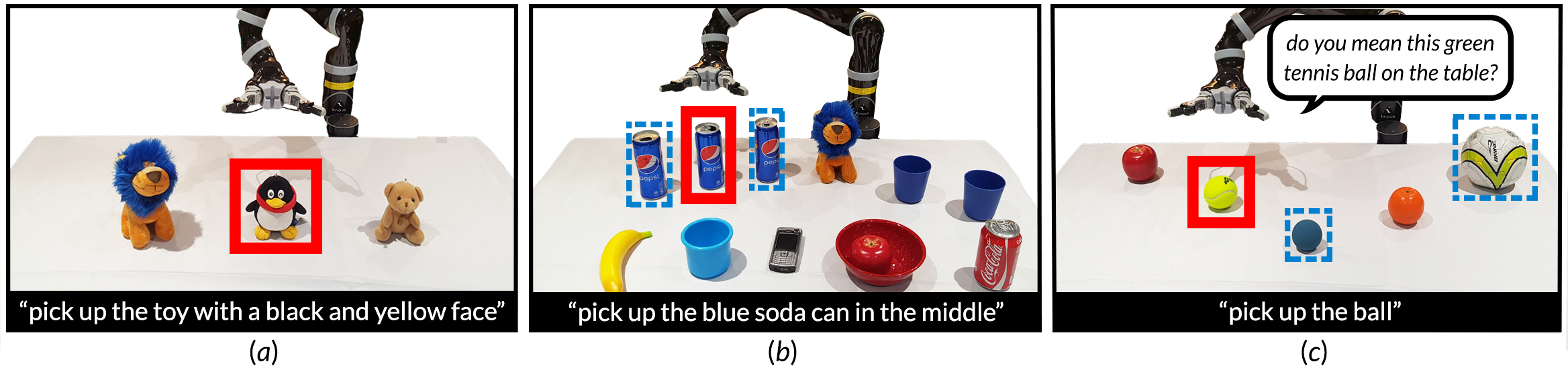}
    \captionsetup{justification=justified,margin=0.1in}
    \captionof{figure}{Interactive visual grounding of referring expressions. 
(\subfig{a}) Ground self-referential expressions. (\subfig{b}) Ground relational expressions. (\subfig{c})~Ask  questions to resolve ambiguity.  Red boxes indicate referred objects. Blue dashed boxes indicate candidate objects. 
See also the accompanying video at \url{http://bit.ly/INGRESSvid}.
}
    
     \label{fig:intro}
    \vspace{8pt}
\end{center}%
}]

\begin{abstract}
This paper presents {\ingress}, a robot system that follows human natural language instructions to pick and place everyday objects. 
The core issue here is the grounding of referring expressions: infer objects and their relationships from input images and  language expressions. 
\ingress allows for unconstrained object categories and unconstrained language expressions. 
Further,  it asks questions to disambiguate referring expressions interactively.
To achieve these, we take the approach of \emph{grounding by generation} and propose  a two-stage neural-network model for grounding. The first stage uses a neural network to generate visual descriptions of objects, compares them with the input language expression, and identifies a  set of candidate objects. The second stage uses another neural network to examine all pairwise relations between the candidates and infers the most likely referred object.
The same neural networks are used for both grounding and question generation for disambiguation. 
Experiments show that \ingress outperformed a state-of-the-art method on the RefCOCO dataset and in robot experiments with humans.
\end{abstract}

\IEEEpeerreviewmaketitle

\section{Introduction}
The human language provides a powerful natural interface for interaction between humans and robots. In this work, we aim to develop a robot system that follows natural language instructions to pick and place everyday objects. 
To do so, the robot and the human must have a shared  understanding of language expressions as well as the environment.  

The core issue here is the \emph{grounding} of natural language referring expressions:   locate objects  from input images and language expressions. To focus on this main issue, we assume for simplicity that the scene is uncluttered and the objects are clearly visible. While prior work on object retrieval typically assumes  predefined object categories, we want to allow for unconstrained object categories so that the robot can handle a wide variety of everyday objects not seen before (Fig.~\ref{fig:intro}). Further,  we want to allow for  rich human language expressions in free form, with no artificial constraints (Fig.~\ref{fig:intro}). Finally, despite the richness of  human language, referring expressions may be ambiguous. The robot should disambiguate such expressions by asking the human questions interactively (Fig.~\ref{fig:intro}). 



To tackle these challenges, we take the approach of \emph{grounding by generation}, analogous to that of analysis by synthesis~\cite{neisser2014cognitive}. We propose a neural-network grounding model, consisting of two networks  trained on large datasets,  to generate language expressions from the input image and compare them with the input referring expression. If the referring expression is ambiguous, the same networks are used to generate questions interactively. We call this approach {\ingress}, for INteractive visual Grounding of Referring ExpreSSions. 

\begin{figure*} 
  \centering
  \includegraphics[width=0.98\textwidth]{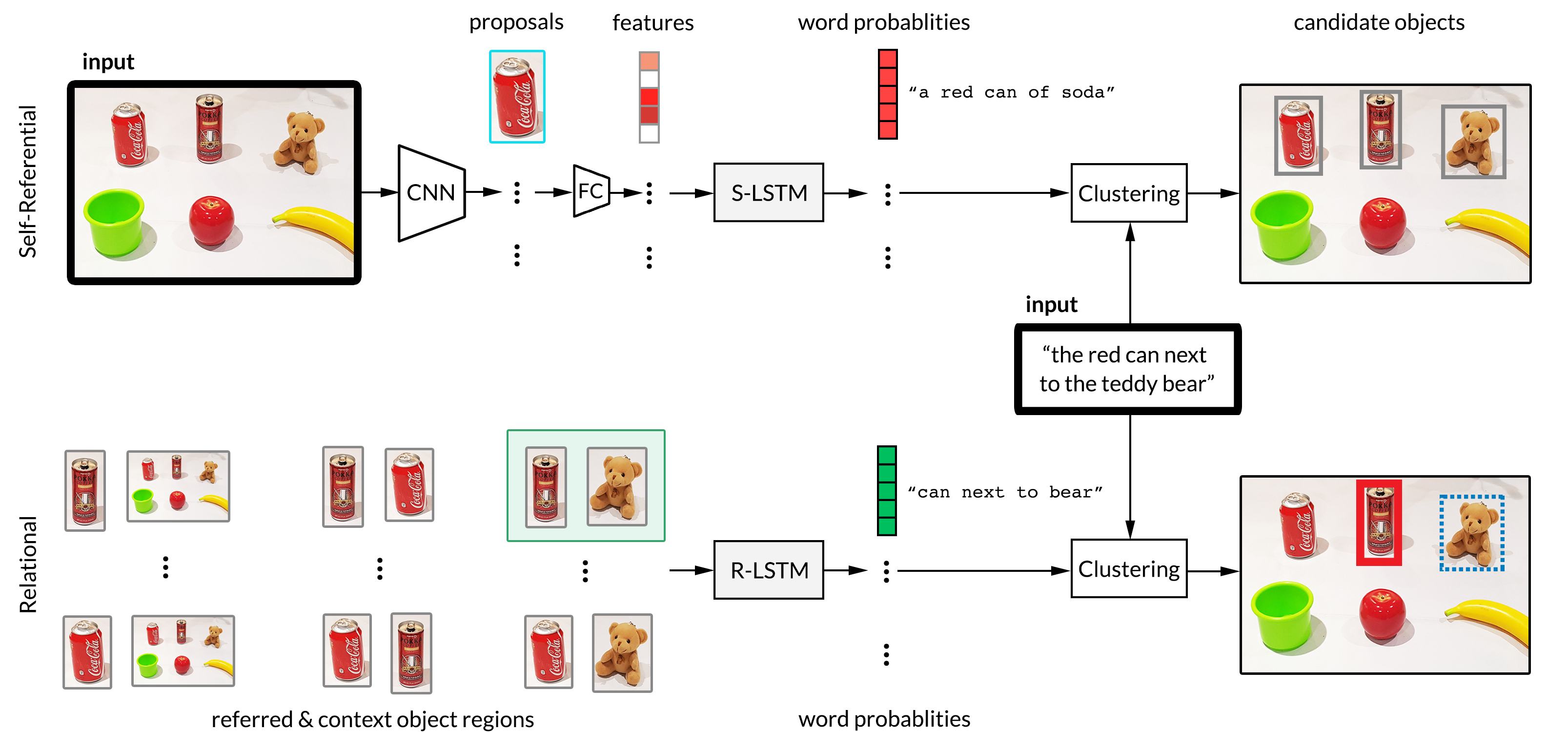}
  \captionsetup{justification=justified,margin=0.1in}
  \caption{ \ingress overview. The first stage grounds  self-referential  expressions and outputs a set of candidate referred objects (top row). The input image goes into a Faster R-CNN~\cite{johnson2016densecap} based localization module to generate image regions representing object proposals. Each image region goes into a fully connected network to extract a feature vector, which in turn  goes into an LSTM network to generate a word probability sequence  that represents an expression distribution describing the image region. The generated expression and the input expression are compared to find candidates for the referred object.   The second stage grounds relational expressions by examining all pairs of candidate image regions (bottom row). Each pair goes into  another LSTM network, which generates a word probability sequence  describing the relation between the pair of image regions. Again, the generated expression and the input expression are compared to find the referred object.}
  
    \label{fig:model_arch}
    \vspace{-10pt}
\end{figure*}

A referring expression may contain \self and \rel sub-expressions. Self-referential expressions describe an object in terms of its own attributes, \eg, name, color, or shape. \relCaps expressions describe an object in relation to other objects, \eg, spatial relations. By exploiting the  compositionality principle of natural language~\cite{werning2012oxford},  
\ingress structurally decomposes the grounding process into two stages (Fig.~\ref{fig:model_arch}). The first stage uses a neural network to ground the \self sub-expressions and identify a set of candidate objects. The second stage uses another neural network to ground   the \rel sub-expressions by examining all pairwise relations between the candidate objects. 
Following earlier 
work~\cite{bisk2016natural,nagaraja2016modeling,tellex2010grounding}, we focus on binary relations here, in particular, visual binary relations. 


We implemented \ingress on a Kinova Mico robot manipulator, with voice input and RGB-D sensing. 
Experiments show that \ingress outperformed a state-of-the-art method  on the RefCOCO test dataset~\cite{kazemzadeh2014referitgame} and in robot experiments with humans. 


\section{Related Work}




Grounding referring expressions is a classic question widely studied  in natural language processing,  computer vision, and robotics (\eg, \cite{ clark1991grounding,pateras1995understanding}). 
A recent  study
identifies four key issues in grounding for human-robot collaborative manipulation:  visual search, spatial reference,  ambiguity, and perspectives (\eg, ``on my left'') \cite{li2016spatial}. Our work addresses the first three issues  and briefly touches on the last one. 



Visual grounding of referring expressions is closely related to object recognition. In robotics, object recognition is often treated as a classification task, with a predefined set of object category labels  \cite{ eppner2016lessons,pangercic2012semantic}. 
These methods  restrict themselves to tasks covered by predefined  visual concepts and simple language expression templates.  
Other methods relax the restriction on language  by developing a joint model of language and perception \cite{fitzgerald2013learning,UW_RSE_ICML2012}, but they have difficulty in scaling up to many different object categories. 

Relations play a critical role in grounding referring expressions for human-robot interaction, as objects are often described in  relation to others. Again, some earlier work treats  relational grounding  
as a classification task with predefined relation templates~\cite{golland2010game,guadarrama2013grounding, huo2016natural}. 
A recent state-of-the-art method performs sophisticated spatial inference on probabilistic models \cite{paul2016efficient}, but  it assumes an explicit semantic map of the world and relies on formal language representation generated by a syntactic parser, which does not account for the visual context and is sensitive to grammatical variations.

Our approach to visual grounding is inspired by recent advances in image caption generation and understanding \cite{hu2016natural, johnson2016densecap, mao2016generation, nagaraja2016modeling, yu2016joint}. By replacing traditional handcrafted visual feature extractors with convolutional neural networks (CNNs) and replacing language parsers with recurrent neural networks (RNNs), these methods learn to generate and comprehend sophisticated human-like object descriptions for unconstrained object categories.
In essence, the networks automatically connect  visual concepts and  language concepts by embedding them jointly in an abstract space. 
Along this line, Nagaraja \etal\ propose a network specifically for grounding relational expressions~\cite{nagaraja2016modeling}.
Similarly, Hu \etal\ propose a modular neural network and train it for grounding end-to-end~\cite{hu2017modeling}. In contrast, we train separate neural networks for self-referential and relational expressions and  use them in a generative manner. This allows us to generate questions for disambiguation, an issue not addressed in these earlier works. 


Ambiguity is an important issue for grounding in practice, but rarely explored. 
The recent work of Hatori \etal\ detects ambiguities, but relies on fixed generic 
question templates, such as ``which one?'', to acquire additional  information for disambiguation~\cite{hatori2017interactively}. \ingress generates object-specific questions, \eg, ``do you mean this blue plastic bottle?''.  

\section{Interactive Visual Grounding}

\subsection{Overview}
\ingress breaks the grounding process into two stages sequentially and trains two separate LSTM networks, \snn and \rnn, for grounding self-referential expressions and relational expressions, respectively (Fig.~\ref{fig:model_arch}). 
The two-stage design takes advantage of the compositionality principle of natural language~\cite{werning2012oxford}; 
it reduces the data requirements for training the networks, as self-referential expressions and relational expressions are  semantically ``orthogonal''. 
Further, the first stage acts as a ``filter'', which significantly reduces the number of candidate objects that must be processed for relational grounding, and improves computational efficiency.  

Each stage  follows the grounding-by-generation approach and uses the LSTM   network to generate a textual description of an input image region or a pair of image regions. It  then compares the generated expression with the input expression to determine the most likely referred object. An alternative is to train the networks directly for grounding instead of generation, but it is then difficult to use them for generating questions in case of ambiguity. 

To resolve ambiguities, \ingress uses  \snn or \rnn  to generate the textual description of a candidate object and fits it to a question template to generate an object-specific question. 
The user  then may provide a correcting response based on the question asked.



\subsection{Grounding \selfCaps Expressions} \label{gng}

Given  an  input image $I$  and  an expression $E$, the first stage of \ingress aims to identify candidate objects from $I$ and  \self sub-expressions of $E$. More formally, let $R$ be a rectangular image region that contains an object. We want to find image regions with high probability $\prob(R \given[\big] E,I)$. Applying the Bayes' rule, we have 
\begin{equation} \label{eq:equality}
\argmax_{R \in \rset}\;\prob(R \given[\big] E,I)
   = \argmax_{R \in \rset} \; p (E \given[\big] R, I)\, \prob(R \given[\big] I), \\
\end{equation}
where \rset\ is the set of all rectangular image regions in $I$. Assuming a uniform prior over the image regions, our objective is then to maximize  $\prob(E \given[\big] R,I)$, in other words, to find an image region $R$ that most likely generates the expression $E$. 
 
To do so, we apply the approach of DenseCap~\cite{johnson2016densecap}, which directly connects image regions that represent object proposals with natural expressions, thus avoiding the need for predefined object  categories. See Fig.~\ref{fig:model_arch} for an overview. 
First,  we use a Faster \mbox{R-CNN} \cite{johnson2016densecap} based localization module to process the input image $I$ and find a set of image regions $R_i, i=1,2, \ldots$, each representing an object proposal. 
We use a fully connected layer to process each region $R_i$ further and produce a  4096-dimensional feature vector $f_i$. 
Next, we feed each feature vector $f_i$ into \snn, an LSTM network,  and predict a sequence $S_i$ of word probability vectors. 
The sequence $S_i$ represents the predicted  expression describing $R_i$. The $j$th vector in $S_i$ represents the $j$th word in the predicted expression, and each element of a vector in $S_i$ gives the probability of a  word. The input sequence $E$ is padded to have the same length as $S$.
We then   calculate the average cross entropy loss (CEL) between  $E$  and $S_i$, or equivalently $\prob(E \mid R_i, I) = \prob(E \mid S_i)$. Effectively, the \snn output allows us to estimate the probability of each word in an expression. The average cross entropy loss over all words in the expression indicates how well it  describes an image region.

Our implementation uses a pre-trained captioning network provided by DenseCap~\cite{johnson2016densecap}. The network was trained on the Visual Genome dataset \cite{krishna2016visual}, which contains around $100,000$ images and  $4,300,000$ expressions, making the model applicable to a diverse range of real-world scenarios.
On  average, each image has $43.5$ region annotation expressions, \eg,  ``cats play with toys hanging from a perch'' and ``woman pouring wine
into a glass''.   

 




\begin{figure}[!t]
  \begin{center}
      \includegraphics[width=0.5\textwidth, center]{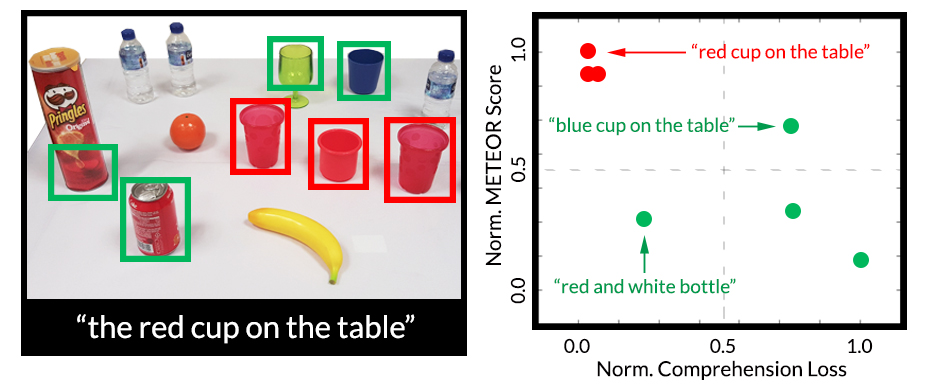}
      \captionsetup{justification=justified, margin=0.1in}
      \caption{Relevancy clustering. 
      Red boxes (left) and red dots (right) indicate relevant objects.
Green boxes and dots indicate irrelevant objects. 
      }
      \label{fig:clustering}
       \vspace{-20pt}
  \end{center}
\end{figure}

\subsection{Relevancy Clustering} \label{clust}

While CEL  measures how well  the input expression  matches the generated sequence of word probability vectors, it is subjected to visual ambiguity as a result of lighting condition variations, sensor noise, object detection failures, etc. Consider the Pringles chip can example in Fig.~\ref{fig:clustering}. The image region contains only part of the can, and it is visually quite similar to a red cup. CEL is thus low, indicating a good fit, unfortunately. Further, the word probability vectors might not consider paraphrases and synonyms, unless explicitly trained with specific examples.

To deal with these issues, we consider an additional  measure, METEOR~\cite{banerjee2005meteor}. 
METEOR is a standard machine translation metric that calculates the normalized semantic similarity score between two sentences. For example, the METEOR score between  ``the green glass'' and ``the green cup'' is $0.83$, and that between ``the green glass'' and ``the blue book'' is $0.06$. METEOR handles paraphrases and synonyms automatically.
We calculate the METEOR measure by generating the most likely expression $E_i$ from  $S_i$ and comparing $E_i$ with the input expression $E$. METEOR, however, has its own limitation. It does not account for the visual context and treats all words in an expression with equal importance. For example, the METEOR score between ``a blue cup on the table'' and ``a red cup on the table'' is high, because most words in the expressions match exactly  (Fig.~\ref{fig:clustering}).

For robustness, we calculate both CEL and METEOR between  $S_i$ and $E$, for $i=1,2, \ldots$\,. 
We then perform $K$-means clustering with normalized CEL \& METEOR values and  choose $K=2$ to form two clusters of relevant and irrelevant  candidate image regions for the referred object (Fig.~\ref{fig:clustering}). Finally,  the relevant cluster  $\rset'$ is sent to the second stage of the grounding model, if $\rset'$ contains multiple candidates.

\subsection{Grounding Relational Expressions} \label{grounding_rel}

In the second stage, we aim to identify the referred object by analyzing its  relations with other objects. We make the usual assumption of  binary  relations~\cite{bisk2016natural,kollar2010toward,nagaraja2016modeling}. While this may appear restrictive, binary relations are  among the most common in everyday expressions. Further, some expressions, such as ``the leftmost cup'', seem to involve complex relations with multiple objects, but it can be, in fact, treated as a binary relation between the referred object and all other objects treated as a single set. Akin to the grounding of \self expressions, we seek a pair of image regions, referred-object region $R$ and context-object region $ R_\mathrm{c}$, with high probability $\prob( R, R_\mathrm{c} \mid E, I)$:

\begin{equation} \label{eq:relationship}
  \argmax_{\shortstack{$\scriptscriptstyle R \in \rset', R_\mathrm{c} \in \rset' \cup \{I\} $\\ $\scriptscriptstyle R\neq R_\mathrm{c} $}} \prob (R, R_\mathrm{c} \given[\big] E, I) = \argmax_{\shortstack{$\scriptscriptstyle R\in \rset', R_\mathrm{c} \in \rset' \cup \{I\} $\\ $\scriptscriptstyle R\neq R_\mathrm{c} $}} \prob (E \given[\big] R, R_\mathrm{c}, I). \\
\end{equation}

Our approach for grounding relational expressions parallels that for grounding self-referential expressions. See Fig.~\ref{fig:model_arch} for an overview.  We form all pair-wise permutations of candidate image regions, including the special one corresponding to the whole image~\cite{mao2016generation}. An image region consists of a feature vector and its bounding box representing its 2D spatial location within the image. We feed all image region pairs into R-LSTM, another LSTM, trained to predict relational expressions. 
By directly connecting image region pairs with  \rel expressions,  we avoid the need for predefined relation templates. For each image-region pair  $(R, R_\mathrm{c})$, we generate the relational expression $E'$. We compute CEL and METEOR between $E'$ and the input expression $E$ over all generated expressions and again perform \mbox{$K$-means} clustering with $K=2$.  If all pairs in the top-scoring cluster contain the same referred object, then  it  is uniquely identified. 
Otherwise, we take all candidate objects to the final disambiguation stage. 

Following the approach of UMD RefExp~\cite{nagaraja2016modeling}, we  trained \rnn  on the RefCOCO training set \cite{kazemzadeh2014referitgame}, which contains around $19,000$ images and $85,000$ referring expressions that describe visual relations between images regions, \eg,  ``bottle on the left''. 
Specifically, we used UMD RefExp's Multi-Instance Learning Negative Bag Margin loss function for training.
We used stochastic gradient decent for optimization, with a learning rate of 0.01 and a batch size of 16. The training converged after $70,000$ iterations and took about a day to train on an Nvidia Titan X GPU. 

\subsection{Resolving Ambiguities} \label{ambiguity}


If the referred object cannot be uniquely identified by grounding the self-referential and relational sub-expressions, the final disambiguation stage of \ingress processes the remaining candidate objects  interactively. For each object, it asks  the human ``Do you mean \ldots?'' and simultaneously, 
commands the robot arm to point to the  location of the object.



Generating object-specific questions is straightforward for \ingress, because of its grounding-by-generation design. To ask a question about an object, we  either use S-LSTM or R-LSTM to generate an expression $E$  and then fit it to the question template ``Do you mean $E$?''
We  start with S-LSTM, as most referring expressions primarily rely  on visual information~\cite{li2016spatial}. 
We generate a self-referential expression for each candidate and check if it is informative.
In our case, an expression $E$ is \textit{informative} if the  average METEOR score between $E$ and all other generated expressions is small, in other words, it is sufficiently different from all other expressions. If the most informative expression has an average METEOR score less than $0.25$, we proceed to ask a question using $E$. Otherwise, we use R-LSTM to generate a relational question.
 
After asking the question, the user can respond ``yes'' to choose the referred object, ``no'' to continue iterating through other possible objects, or provide a specific correcting response to the question, \eg,  ``no, the cup on the left''. To process the correcting response, we re-run \ingress with the identified candidate objects and the new expression.

\section{System Implementation}

\begin{figure}[!t]
  \begin{center}
      \includegraphics[width=0.48\textwidth, center]{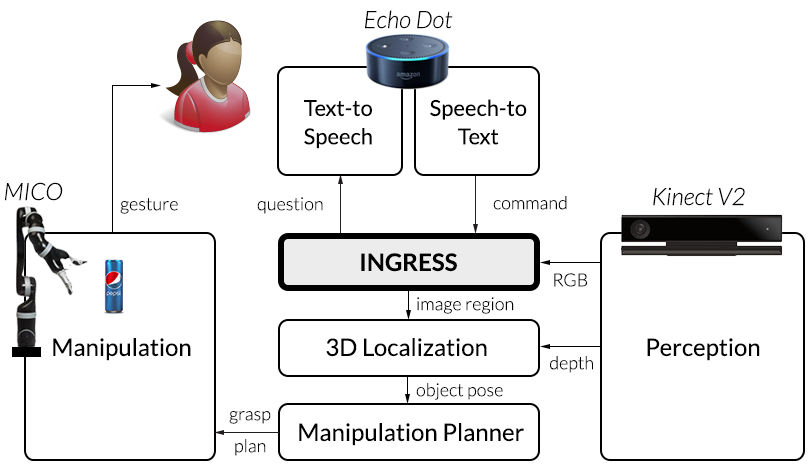}
      \captionsetup{justification=justified}
      \caption{An overview of the system architecture.}
      \label{fig:sys_arch}
       \vspace{-20pt}
  \end{center}
\end{figure}

To evaluate our approach, we implemented \ingress on a robot manipulator, with voice input and RGB-D sensing.  Below we briefly describe the system setup (Fig. \ref{fig:sys_arch}). 

\subsection{Visual Perception and Speech Recognition}
Our grounding model  takes in as input  an RGB image and a textual referring expression, and outputs a 2D bounding box containing the referred object in the image (Fig. \ref{fig:model_arch}).
Our system uses a Kinect2 RGB-D camera for visual perception and an Amazon Echo Dot device to synthesize  the referring expression from voice input. 

\subsection{Grounding Networks}
The localization module for object detection uses a non-maximum suppression threshold of $0.7$ and a final output threshold of $0.05$ for minimal overlap between bounding boxes in uncluttered scenes. 
 
\snn and \rnn have a vocabulary size of  $10,497$ and $2,020$, respectively. The maximum sequence length for both is 15 words. 


\subsection{Object Manipulation}
Our system uses a 6-DOF Kinova MICO arm for object manipulation. It is currently capable of  two high-level actions, \cmd{PickUp} and \cmd{PutIt}.  For \cmd{PickUp}, the system first uses the Kinect2 depth data corresponding to the selected 2D bounding box  and localizes the referred object in 3D space. 
It then  plans a  top-down or a side grasp pose based  on the object size, as well as a path to reach the pose. For \cmd{PutIt}, the system similarly identifies the placement location.   
It moves the end-effector to  position it directly above the desired location and then simply opens up the gripper. This simple set up is sufficient for our experiments. However, we plan to  integrate state-of-the-art methods for grasping and manipulating novel objects~\cite{mahler2017dex}.

 
\subsection{Software and Hardware Platform}
The entire system~(Fig. \ref{fig:sys_arch}), with components for RGB-D visual perception, grounding, and manipulation planning, is implemented under the  Robot Operating System (ROS) framework and runs on a PC workstation with an Intel i7 Quad Core CPU and an NVIDIA Titan X GPU. 
The  grounding model runs on the GPU. 

\subsection{Perspective Correction}

Referring expressions are conditioned on perspectives~\cite{li2016spatial}:
object-centric (\eg, ``the bottle next to the teddy bear''), user-centric (\eg, ``the bottle on my left''), or robot-centric (\eg, ``the bottle on your right''). 
Object-centric expressions are  handled directly by the grounding model. User-centric and robot-centric expressions require special treatment. Handling perspectives reliably is a complex issue.  Here we provide a  solution  dealing with the simple, common cases in a limited way.  Given two detected viewpoints for the user and the robot perspective, the system associates a set of possessive keywords such as ``my'', ``your'', \etc\ with each viewpoint. It then  matches the input expression against the keyword list to select a viewpoint and performs a corresponding geometric transformation of generated bounding boxes to the specified viewpoint frame.

For the geometric transformation, we first compute a 3D centroid for each bounding box using the depth data. The centroid is then projected onto the image plane of either the robot's or the user's viewpoint. This projected point is taken to be the center of the new bounding box. The size of the box is then scaled linearly with respect to the distance between the centroid and the viewpoint while the original aspect ratio is maintained.


\section{Experiments}
We evaluated our system under three settings. First, we  evaluated for grounding accuracy and generalization to  a wide variety of objects and relations on the RefCOCO dataset~\cite{kazemzadeh2014referitgame}.
Next, we evaluated for  generalization  to unconstrained language expressions in robot experiments with humans.
In both cases, \ingress outperformed UMD Refexp~\cite{nagaraja2016modeling}, a state-of-the-art method in visual grounding.
Finally, we  evaluated  for  effectiveness of disambiguation and found that \ingress, through object-specific questions,  sped up task completion by 
$1.6$ times on  average. 

In  uncluttered scenes with 10--20 objects,  the overall voice-to-action cycle takes 2--5 seconds for voice-to-text synthesis,  retrieving the synthesized text from Amazon's service, grounding, visual perception processing, and manipulation planning  for picking or putting actions by the 6-DOF robot arm. In particular, 
 grounding takes approximately $0.15$ seconds. 
 

\subsection{RefCOCO Benchmark}

\newcolumntype{P}[1]{>{\centering\arraybackslash}p{#1}}
\setlength\extrarowheight{4.5pt}

\begin{table}[t]
\centering
\label{my-label}
\captionsetup{justification=justified,margin=0.1in}
\caption{Grounding accuracy  of UMD Refexp and \ingress on the RefCOCO dataset, with human-annotated ground-truth (HGT) object proposals and automatically generated MCG object proposals.}
\begin{tabular}{c@{\hspace*{16pt}}c@{\hspace*{6pt}}c@{\hspace*{16pt}}c@{\hspace*{6pt}}c}
\hline

\hline
Dataset      & \multicolumn{2}{c}{HGT  (\%)} & \multicolumn{2}{c}{MCG (\%)} \\ 
 & UMD Refexp  & INGRESS        & UMD Refexp    & INGRESS       \\ \hline
Val           & 75.5        & \textbf{77.0}  & 56.5          & \textbf{58.3} \\
TestA         & 74.1        & \textbf{76.7}  & 57.9          & \textbf{60.3} \\
TestB         & 76.8        & \textbf{77.7}  & \textbf{55.3} & 55.0          \\ \hline
\end{tabular}
\vspace{-2pt}
\label{results}
\end{table}

The RefCOCO dataset contains images and corresponding referring expressions, which   use both \self and \rel information to uniquely identify objects in images. 
The dataset covers a wide variety of different objects and is well suited for evaluating generalization  to unconstrained object categories.
Our evaluation measures the accuracy at which a model can locate an image region, represented as an image bounding box, given an expression describing  it unambiguously.   

We compared \ingress with UMD Refexp \cite{nagaraja2016modeling}
on the \mbox{RefCOCO} dataset. UMD Refexp's approach to relational grounding is similar to that of \ingress (see Section \ref{grounding_rel}), but there are two key differences. First, UMD Refexp uses feature vectors from an image-net pre-trained VGG-16 network, whereas \ingress uses captioning-trained feature vectors from the self-referential grounding stage. Second, for images with more than 10 object proposals, UMD Refexp randomly samples 9 candidates for \rel grounding, while  \ingress only examines the pairs of objects proposals chosen by 
the \self grounding stage.

\paragraph{Procedure} 
The RefCOCO dataset consists of a training set, a validation set (Val), and two test sets (TestA and Test B).
TestA contains images with multiple people. TestB contains images with multiple instances of all other objects.
TestA contains 750 images with 5657 expressions. TestB contains 750 images with 5095 expressions. 
Val contains 1500 images with 10834 expressions.
Following UMD Refexp, we use  both human-annotated ground-truth object proposals and automatically generated MCG proposals \cite{arbelaez2014multiscale} in our evaluation.


\paragraph{Results}  
The results are reported in Table~\ref{results}.
The correctness of a grounding result is based on  the overlap between the output  and the ground-truth image regions. The grounding is deemed correct if the  intersection-over-union (IoU) measure between the two region is greater than $0.5$.
Table~\ref{results} shows that \ingress outperforms UMD Refexp in most cases, 
but the improvements are small.
\ingress adopts a two-stage grounding process in order to reduce the number of relevant object proposals processed in complex scenes. On average, the validation and test sets contain 10.2 ground-truth object proposals and 7.4 MCG object proposals per image. As the number of object proposals per image is small, the two-stage grounding process does not offer significant benefits.  


We also observed that images containing people have greater improvement in accuracy than those containing only objects. 
This likely results from the large bias in the number of images containing people in the Visual Genome dataset~\cite{krishna2016visual}. Future work may build a more balanced dataset with a greater variety of common objects for training the grounding model. 

\begin{figure}[!t] 
\begin{center}
      \includegraphics[trim={0, 0cm, 0cm, 2cm},width=0.47\textwidth, center]{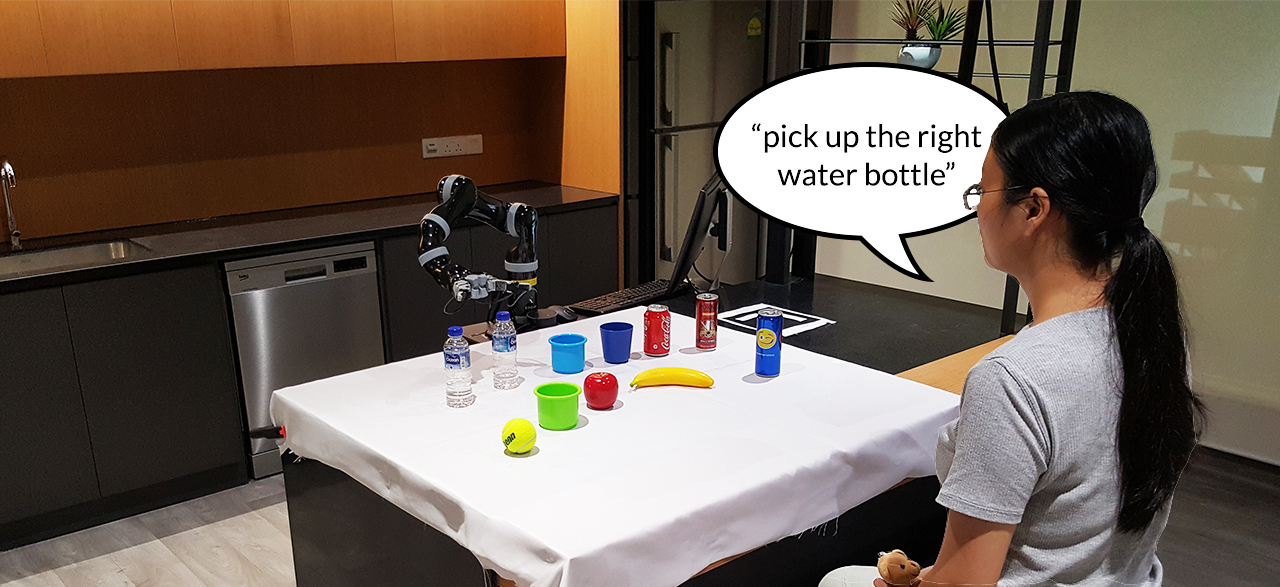}
      \captionsetup{justification=justified}
      \caption{Experimental setup for robot experiments.}
      \label{fig:exp_setup}
  \end{center}
  \vspace{-15pt} 
\end{figure}


\subsection{Robot Experiments} \label{collab_mani}

We also  assessed the performance of our grounding model in a realistic human-robot collaboration context and particularly, to study its ability in handling unconstrained language expressions. In our experiments, a group of participants provided natural language instructions to a 6-DOF manipulator to pick and place objects (Fig.~\ref{fig:exp_setup}).

Again, we compared \ingress with UMD Refexp~\cite{nagaraja2016modeling}. 
We also conducted an ablation study, which compared pure \self grounding (\ingressS) and the complete model with both \self and \rel grounding. For \ingressS, we directly used the image region with the lowest cross-entropy loss from the \self stage. For \ingressSR, we used the region chosen by the full model. Further,  both \ingressS and \ingressSR, used the object proposals  generated by the \self stage, whereas UMD Refexp used MCG proposals~\cite{arbelaez2014multiscale}. 
All methods used a large number of object proposals. So the probability of randomly picking the referred object was very low.

\paragraph{Procedure} 
Our study involved 16 participants (6 female, 10 male) recruited from a university community. All subjects were competent in spoken English. Each participant was shown 15 different scenarios with various household objects arranged in an uncluttered manner. 

In each scenario, the experimenter asked the participant to describe a specific object to the robot. The experimenter gestured at the object without hinting any language descriptions. Before instructing the robot, the subjects were given three generic guidelines: the object description has to be simple, unique (unambiguous), and any perspectives taken should be stated explicitly \eg, `my left', `your right'. Although, these guidelines were not strictly enforced. 
Upon receiving an expression, all 3 models (\ingressS, \ingressSR, UMD Refexp) received the same image and expression as input, and 3 trials were run simultaneously. A trial was considered successful if the robot located the specified object on its first attempt. 


The average number of objects per scenario was 8.  
And the maximum number of identical objects was 3. The scenarios were carefully designed such that 66\% required \rel cues, 33\% involved perspective taking, and 100\% required \self information. For assessing perspectives, the participant was positioned at one of the four positions around the robot: front, behind, left, right.  Also, since the models were trained on public datasets, all objects used in the experiments were `unseen'. However, generic objects like apples and oranges had minimal visual differences to the training examples. 

\begin{figure}[!t]
  \begin{center}
  \hspace*{0.55cm}      \includegraphics[trim={0, 1cm, 0cm, 2.2cm},width=0.43\textwidth, center]{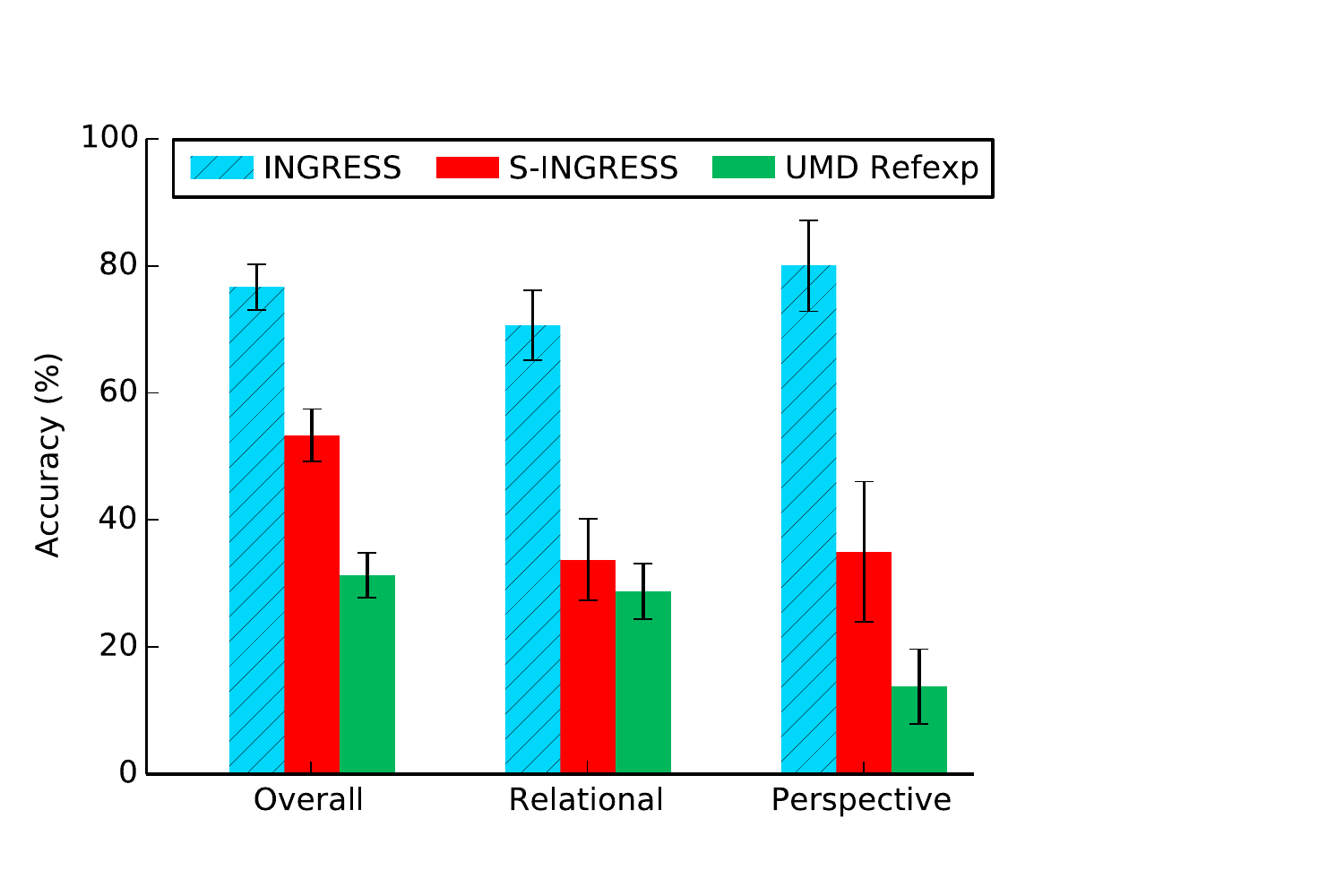}
      \captionsetup{justification=justified}
      \caption{Grounding accuracy in robot experiments with humans. Error bars indicate 95\% confidence intervals.}
      \label{fig:barplot}
  \end{center}
  \vspace{-20pt}
\end{figure}


\paragraph{Results} The results (Fig. \ref{fig:barplot}) show that overall \ingressSR significantly outperforms both \ingressS ($p < 0.001$ by t-test) and UMD Refexp ($p < 0.001$ by t-test). 
\ingressS is effective in locating objects based on \self information. However, it fails to infer relationships, as each image region is  processed in isolation.
While UMD Refexp in principle makes use of both \self and \rel information, it performs poorly in real-robot experiments, particularly, in grounding \self expressions. 
UMD Refexp is trained on a relatively small dataset, RefCOCO, with mostly \rel expressions. Its ability in grounding \self expressions is inferior to that of  \ingressSR and \ingressS.
Further, \ingress uses relevancy clustering to narrow down a set of object proposals for relationship grounding, whereas UMD Refexp examines a randomly sampled subset of object proposal pairs, resulting in increased errors. 
Finally, UMD Refexp is incapable of handling perspectives, as it is trained on single images without viewpoint information. 

During the experiments, we  observed that referring expressions varied significantly across participants. 
Even a simple object such as an apple was described in many different ways as ``the red object'', ``the round object'', ``apple in middle'', ``fruit'' \etc\, Likewise, relationships were also described in many different variations,  \eg, ``the can in the middle'', ``the second can'', \etc\, Our model correctly handled most of these variations. 

Occasionally, participants used complex ordinality constraints, \eg, ``the second can from the right on the top row''. None of the models examined here, including \ingress, can handle ordinality constraints. Other common failures include text labels and brand names on objects, \eg, ``Pepsi''. 


\subsection{Disambiguation} \label{asking_ques_eval}
We conducted a user study to examine the effectiveness of \ingress in asking disambiguating questions.
\ingress asks object-specific questions  (\eg, ``do you mean this red cup?''), and the user may provide a correcting response (\eg, ``no, the red cup on the right''). We compared with a baseline method similar to the work of \citet{hatori2017interactively}. 
There, the robot exhaustively points at objects while asking a generic question ``do you mean this object?'', and expects a yes/no answer from the user.
Specifically, we examined two issues: 
\begin{itemize}
\item Does  \ingress' approach of asking object-specific questions improve grounding in terms of the time required to resolve ambiguities? 
\item Are the generated questions effective in communicating the required additional information from the user?  
\end{itemize}

\begin{figure}[!t] 
  \begin{center}
   \includegraphics[trim={0, 0.4cm, 0cm, 0cm},width=0.5\textwidth, center]{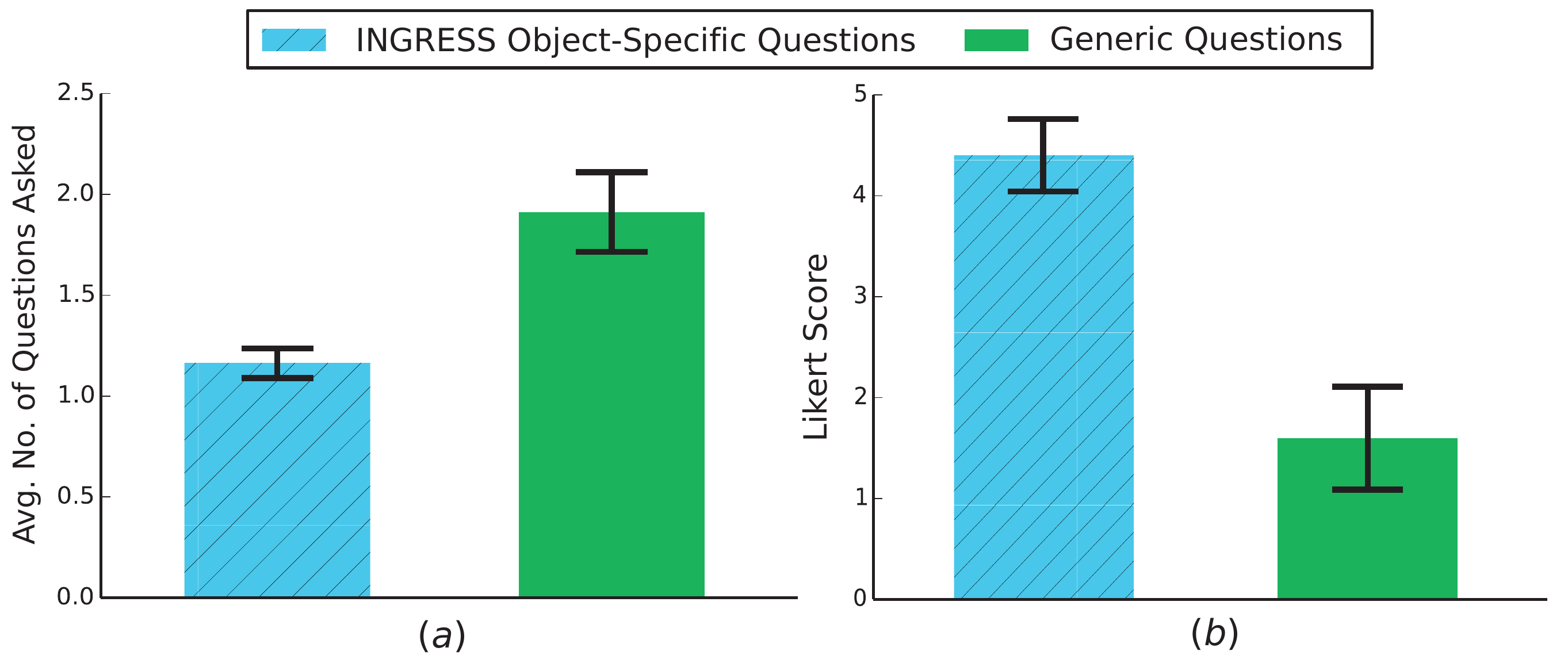}
      \captionsetup{justification=justified}
      \caption{Disambiguation performance. Error bars indicate 95\% confidence intervals. (\subfig{a}) Average number of disambiguation questions asked. (\subfig{b})  User survey on the robot's effectiveness in communicating the additional information required for disambiguation.}
      \label{fig:ask_result}	
      
  \end{center}
  \vspace{-15pt}
\end{figure}

\begin{figure*}[!t]
  \centering
  \includegraphics[width=0.96\textwidth]{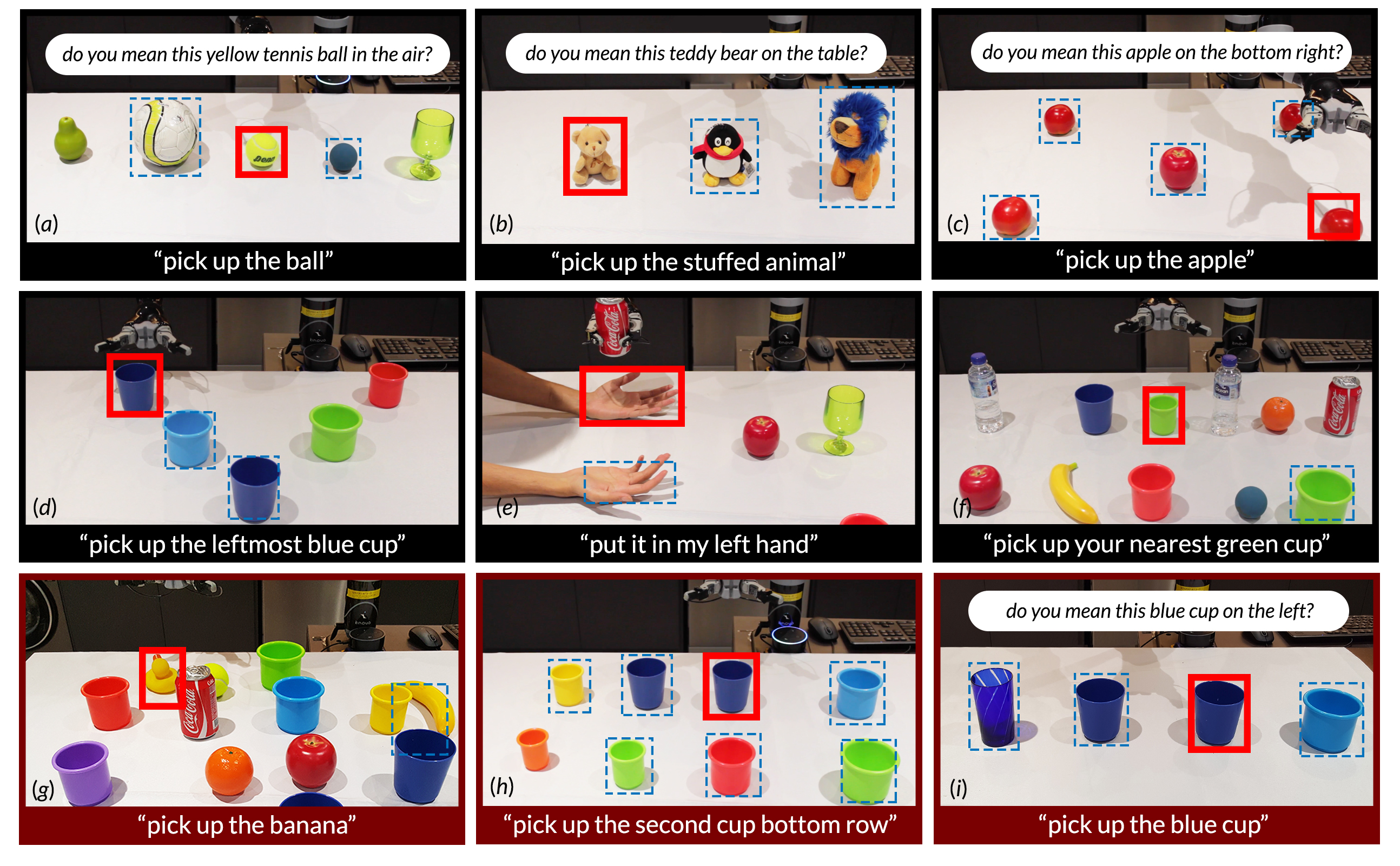}
  \captionsetup{margin=0.6cm}
    \caption{A sample of interactive grounding results. Red boxes indicate the objects chosen by  \ingress. Blue dashed boxes indicate candidate objects. The first two rows show successful results and disambiguation questions. The last row shows some failure cases.}
    \vspace{5pt}
    \label{fig:qualitative}
    \vspace{-10pt} 
\end{figure*} 

\paragraph{Procedure} The study was conducted with the same 16 participants from Section~\ref{collab_mani}. 8 participants for the baseline condition, and 8 participants for our method. Each subject was shown 10 different scenarios with various household object. For each scenario, the experimenter initiated the trial by giving the robot an ambiguous instruction \eg, ``pick up the cup'' in scene with a red cup, blue cup, green cup and yellow cup. The robot chose one of the candidate objects, and asked a question. Then the participant was asked to choose another potential objects, which was not chosen by robot, and had to correct the robot to pick that object. For the baseline, the participants could only use yes/no corrections. For our method, they could correct the ambiguous expression with additional information \eg, ``no, the red cup'' or ``no, the cup on the left''. 

Half of the scenarios were visually ambiguous, and the other half were relationally ambiguous. The average number of ambiguous objects per scenario was 3, and the maximum was 7. We conducted a total of 160 trials. 
In all trials, the participants were eventually able to correct the robot to find the required object.

\paragraph{Results} Fig.~\ref{fig:ask_result}{\subfig{a}} shows that  \ingress (average 1.16 questions) is more efficient in disambiguation than   the baseline method (average 1.91 questions), with $p < 0.001$ by the t-test. 
While the difference appears small, it is statistically significant. Further,  there are typically only 2--4 objects involved in the disambiguation stage. The improvement is thus practically meaningful as well. 

We also conducted a post-experiment survey and asked participants to rate the agreement question ``the robot is effective in communicating what additional information is required for disambiguation'' on a 5-point Likert scale.  
Again, \ingress scores much higher than the baseline method, $4.4$ versus $1.6$ with a significance of $p < 0.001$ by the Kruskal-Wallis test (Fig.~\ref{fig:ask_result}{\subfig{b}}).

During the experiments, we observed that participants often mimicked the language that the robot used.   
On  average, approximately 79\% of the correcting responses mirrored the robot's questions. For example, when robot asks ``do you mean this apple on the \textit{bottom right}?'', the  user responds  ``no, the apple on the \textit{top left}''.  A few participants also commented that they would not  have  used certain descriptions, \eg, ``top left'', if it were not for the robot's question. This is consistent with the psycholinguistic phenomenon of \textit{linguistic accommodation}~\cite{gallois2015communication}, in which participants in a conversation adjust their language style and converge to a common one. It is interesting to observe here that linguistic accommodation occurs not only between humans and humans, but also between humans and robots. Future works could study this in more detail.

\subsection{Examples}

Fig.~\ref{fig:qualitative} shows a  sample of interactive grounding results. Fig.~\ref{fig:qualitative}(\subfig{a}--\subfig{b})  highlight rich  questions generated by \ingress.
The questions are  generally clear and discriminative, though occasionally they contain  artifacts, \eg, ``ball in the air'' due to biases in the training dataset.
Although our system is restricted to binary relations,  Fig.~\ref{fig:qualitative}(\subfig{c}--\subfig{d}) show some  scenes that contain complex, seemingly non-binary relationships. The referred apple is at the bottom right corner of the entire image, treated as a single object. Likewise, the selected blue cup is the closest one to the left edge of the image. 
Fig.~\ref{fig:qualitative}(\subfig{e}--\subfig{f}) showcase user-centric and robot-centric perspective corrections, respectively. They enable users to adopt intuitive viewpoints such as ``my left''.
Fig.~\ref{fig:qualitative}(\subfig{g}--\subfig{i}) show  some common failures.
\ingress  has difficulty with cluttered environments. Partially occluded objects,  such as Fig.~\ref{fig:qualitative}(\subfig{g}),  often result in false positives.
It also cannot handle complex relationships, such as Fig.~\ref{fig:qualitative}(\subfig{h}), which requires counting (``third'') or grouping objects (``row'', ``all four''). 
Fig.~\ref{fig:qualitative}(\subfig i) is an interesting case. The user's intended object is the second cup from the left, but the input expression is ambiguous. While the generated question is not discriminative,  the robot arm's pointing gesture helps to identify the correct object after two questions.


\section{Discussion}
\label{sec:discussion}
While the experimental results are very promising, \ingress has several limitations (see Fig. \ref{fig:qualitative}). 
First, it handles only binary relations between the referred and context objects. It is not easy to scale up the network to handle truly tertiary or more complex relations. Recent work on relational networks~\cite{santoroNIPS2017} trained on complex relationship corpora~\cite{johnson2017clevr, suhr2017corpus} may help.  
Further, integrating non-verbal cues such as gestures and gaze \cite{palinko2016robot, fischer2016markerless} may reduce the need for interpreting complex instructions.
Second, \ingress  relies on keyword matching to understand perspectives. 
Augmenting the training set with perspective-bearing expressions could allow the system to generalize better.
Third, the  clustering components of the grounding model are currently hard-coded. If we represent them as neural network modules, the grouping of relevant objects can be learned simultaneously with other components. 
Lastly, \ingress cannot handle cluttered environments with partially occluded objects. Systematically moving away objects to reduce uncertainty \cite{li2016act} may help.




\section{Conclusion} \label{sec:conclusion}
We have presented \ingress, a neural network model for grounding unconstrained natural language referring expressions. By training the network on large datasets, \ingress handles an unconstrained, wide variety of everyday objects. In case of ambiguity, \ingress is capable of asking object-specific disambiguating questions. The system outperformed UMD Refexp substantially in robot experiments with humans and generated interesting interactions for disambiguation of referring expressions. 
Even though we are far from achieving a perfect shared understanding of the world between humans and robots, we hope that our work is a step in this direction. It points to several important, exciting issues (Section~\ref{sec:discussion}), which will be our immediate next steps. An equally important, but different direction is the grounding of verbs~\cite{kollar2010toward} to expand the repertoire of robot actions.

\section*{Acknowledgments}
We thank members of the Adaptive Computing Lab at NUS for thoughtful discussions. We also thank the anonymous reviewers for their careful reading of the manuscript and many suggestions that have helped to improve the paper. This work was supported by the NUS School of Computing Strategic Initiatives.

\balance
\bibliographystyle{abbrvnat}
\bibliography{references}



    


    

\end{document}